\documentclass{article}
\usepackage[table]{xcolor}
\usepackage{PRIMEarxiv}
\usepackage[utf8]{inputenc}
\usepackage[T1]{fontenc}
\usepackage[hyphens]{url}
\usepackage[colorlinks=true,breaklinks=true]{hyperref}
\usepackage{url}
\usepackage{booktabs}
\usepackage{amsfonts,amsmath,amssymb}
\usepackage{nicefrac}
\usepackage{microtype}
\usepackage{fancyhdr}
\usepackage{graphicx}
\graphicspath{{media/}}
\usepackage{algorithmic}
\usepackage{textcomp}
\usepackage{tikz}
\usepackage{colortbl}
\usepackage{multirow}
\usepackage{array}
\usepackage{siunitx}
\usepackage{pdflscape}
\usepackage{caption}
\usepackage{longtable}
\usepackage{tcolorbox}
\tcbuselibrary{skins}
\usepackage{pgfplots}
\pgfplotsset{compat=1.18} 
\usetikzlibrary{arrows.meta, positioning, shapes.misc}
\usepackage{pifont}
\usepackage[hyphens]{url}
\usepackage[toc,page]{appendix}

\hypersetup{
    colorlinks=true,
    linkcolor=blue,
    citecolor=blue,
    urlcolor=magenta,
    pdftitle={Performance of Large Language Models in Supporting Medical Diagnosis and Treatment},
    pdfauthor={Diogo Sousa, Guilherme Barbosa, Catarina Rocha, Dulce Oliveira},
    pdfkeywords={Artificial Intelligence in Healthcare, Clinical Decision Support, Large Language Models, Medical Diagnosis, Natural Language Processing}
}
\newcommand{\goldmodel}[1]{\colorbox{yellow!60!orange!30}{\strut #1}}
\newcommand{\silvermodel}[1]{\colorbox{gray!25!white!80!black}{\strut #1}}
\newcommand{\bronzemodel}[1]{\colorbox{orange!50!brown!30}{\strut #1}}

\definecolor{lightgray}{RGB}{245,245,245}
\definecolor{highlightrow}{RGB}{255, 247, 240}
\definecolor{abovebench}{RGB}{213,232,212}
\definecolor{belowbench}{RGB}{255,224,224}
\definecolor{googleblue}{HTML}{BBDEFB}
\definecolor{openaigreen}{HTML}{C8E6C9}
\definecolor{anthropicpurple}{HTML}{D1C4E9}
\definecolor{deepseekorange}{HTML}{FFE0B2}
\definecolor{metared}{HTML}{FFCDD2}
\definecolor{xaicyan}{HTML}{B2EBF2}
\definecolor{alibabayellow}{HTML}{FFF9C4}
\definecolor{mistralteal}{HTML}{B2DFDB}
\definecolor{clow}{RGB}{0, 102, 204}
\definecolor{cmed}{RGB}{255, 191, 15}
\definecolor{chigh}{RGB}{220, 50, 47}
\definecolor{contamLowColor}{RGB}{104,198,104} 
\definecolor{contamMedColor}{RGB}{255, 215, 0}   
\definecolor{contamHighColor}{RGB}{220, 50, 47}
\usepackage{pgfplots}
\pgfplotsset{compat=1.18}
\usetikzlibrary{arrows.meta, positioning, shapes.misc, patterns, backgrounds} 

\definecolor{lightgray}{RGB}{245,245,245}
\definecolor{highlightrow}{RGB}{255, 247, 240} 
\definecolor{abovebench}{RGB}{213,232,212}     
\definecolor{belowbench}{RGB}{255,224,224}     
\definecolor{exceedsmax}{RGB}{255,215,0}       

\definecolor{googleblue}{HTML}{BBDEFB}

% ...

\newtcolorbox{perfbox}[1]{ 
    colback=#1!20, 
    colframe=#1!80!black,
    boxrule=0.5pt,
    arc=2mm, 
    left=2mm, right=2mm, top=1mm, bottom=1mm,
    boxsep=1mm,
    sharp corners=all, 
    nobeforeafter, 
    before upper={\parindent0pt} 
}
\definecolor{perfbest}{RGB}{0,120,210}     
\definecolor{perfmostcorrect}{RGB}{24,140,40} 
\definecolor{perfmostwrong}{RGB}{255,150,22}  
\newcommand{\GoldMedal}{\textcolor{orange!80!yellow}{\ding{192}}}   
\newcommand{\SilverMedal}{\textcolor{gray}{\ding{193}}}             
\newcommand{\BronzeMedal}{\textcolor{orange!70!brown}{\ding{194}}}  
\newcommand{\MaxScore}{65.61}
\newcommand{\contamLow}{\textcolor{contamLowColor}{\textbullet}\hspace*{1.5pt}}
\newcommand{\contamMed}{\textcolor{contamMedColor}{\textbullet\hspace*{0.5pt}\textbullet}\hspace*{1.5pt}}
\newcommand{\contamHigh}{\textcolor{contamHighColor}{\textbullet\hspace*{0.5pt}\textbullet\hspace*{0.5pt}\textbullet}}

\newtcolorbox{sidebox}[2][black!12]{enhanced, colback=#1, colframe=black!80!black,
  boxrule=0pt, left=1em, right=1em,
  toptitle=1mm, bottomtitle=1mm,
  before upper={\strut},
  borderline west={2.5pt}{0pt}{black!75},
  sharp corners, boxsep=1.2ex,
  fonttitle=\bfseries, title={#2},
  before skip=1em, after skip=1em}

\title{Performance of Large Language Models in Supporting Medical Diagnosis and Treatment}
\author{
  Diogo Sousa \\
  Biomechanics and Health Unit \\
  FEUP - Faculty of Engineering of the University of Porto \\
  INEGI – Institute of Science and Innovation \\
  in Mechanical and Industrial Engineering \\
  Porto, Portugal\\
  \texttt{dssousa@inegi.up.pt} \\
  \And
  Guilherme Barbosa \\
  Biomechanics and Health Unit \\
  INEGI – Institute of Science and Innovation \\
  in Mechanical and Industrial Engineering \\
  Porto, Portugal\\
  \texttt{gbarbosa@inegi.up.pt} \\
  \And
  Catarina Rocha \\
  Biomechanics and Health Unit \\
  FEUP - Faculty of Engineering of the University of Porto \\
  INEGI – Institute of Science and Innovation \\
  in Mechanical and Industrial Engineering \\
  Porto, Portugal\\
  \texttt{cmrocha@inegi.up.pt} \\
  \And
  Dulce Oliveira \\
  Biomechanics and Health Unit \\
  INEGI – Institute of Science and Innovation \\
  in Mechanical and Industrial Engineering \\
  Porto, Portugal\\
  \texttt{doliveira@inegi.up.pt} \\
}
\begin{document}
\maketitle

\begin{abstract}
The integration of Large Language Models (LLMs) into healthcare holds significant potential to enhance diagnostic accuracy and support medical treatment planning. These AI-driven systems can analyze vast datasets, assisting clinicians in identifying diseases, recommending treatments, and predicting patient outcomes. This study evaluates the performance of a range of contemporary LLMs, including both open-source and closed-source models, on the \textbf{2024 Portuguese National Exam for medical specialty access (PNA)}, a standardized medical knowledge assessment. Our results highlight considerable variation in accuracy and cost-effectiveness, with several models demonstrating performance exceeding human benchmarks for medical students on this specific task. We identify leading models based on a combined score of accuracy and cost, discuss the implications of reasoning methodologies like Chain-of-Thought, and underscore the potential for LLMs to function as valuable complementary tools aiding medical professionals in complex clinical decision-making.

\vspace{1pt} 
\noindent
\textbf{Keywords:} \small
Artificial Intelligence in Healthcare, Clinical Decision Support, Large Language Models, Medical Diagnosis, Natural Language Processing

\end{abstract}
\section{Introduction}
\label{sec:introduction}

Recent advancements in artificial intelligence (AI), particularly Large Language Models (LLMs) such as those developed by Google (Gemini), OpenAI (GPT series), Anthropic (Claude), Meta (LLaMA), and others, have significantly enhanced the potential for supporting medical diagnostics and treatment decision-making. These models, characterized by their ability to process and generate human-like text, show promise in revolutionizing aspects of clinical workflows, patient education, and medical research \cite{b1}. Previous studies have explored the performance of earlier LLMs, such as ChatGPT, on medical licensing exams, including prior editions of the Portuguese National Residency Access Examination (PNA) \cite{b2}, demonstrating promising capabilities but often falling short of top human performance thresholds consistently across all specialties.

\begin{figure}[h!]
\centering
\includegraphics[width=0.7\linewidth]{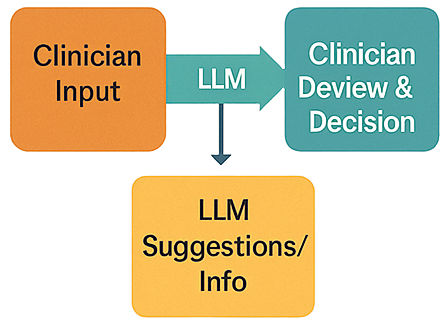}
\caption{Conceptual Diagram}
\label{fig:workflow_diagram}
\end{figure}

In many clinical scenarios, particularly those involving complex differential diagnoses or treatment planning, rapid access to and synthesis of relevant information is critical. This study explores the capabilities of current LLMs in the medical domain, focusing on their potential to serve as \textit{adjunctive tools} for healthcare professionals. The objective is not to replace clinical judgment but to investigate how LLMs can assist doctors by processing complex information, suggesting potential diagnostic pathways, and aiding in treatment considerations. Figure \ref{fig:workflow_diagram} represents this approach. We evaluate a diverse set of both freely available (open-source) and commercially licensed (closed-source) models to understand the performance landscape across different access paradigms, balancing cutting-edge capabilities often found in closed models with the transparency and customizability offered by open-source alternatives. We evaluate models based not only on accuracy but also cost-effectiveness—an essential consideration in healthcare, where budgets and resource allocation can strongly influence the adoption of new technologies. By examining their accuracy on a challenging medical knowledge assessment, we aim to provide insights into their current strengths and limitations as potential aids in the diagnostic and treatment process.

\section{The Evolution of LLMs in Healthcare}
\label{sec:evolution}

Initially, LLMs faced considerable limitations when dealing with the nuances of medical jargon and the complexity of clinical reasoning. However, contemporary models have made dramatic improvements, driven by greater computational power, larger and more diverse training datasets, and advanced training techniques, including reinforcement learning from human feedback (RLHF) applied to CoT.

\subsection{Advances in Foundational Architectures}
\label{subsec:architectures}
\begin{figure}[h!]
    \centering
    \includegraphics[width=0.75\linewidth]{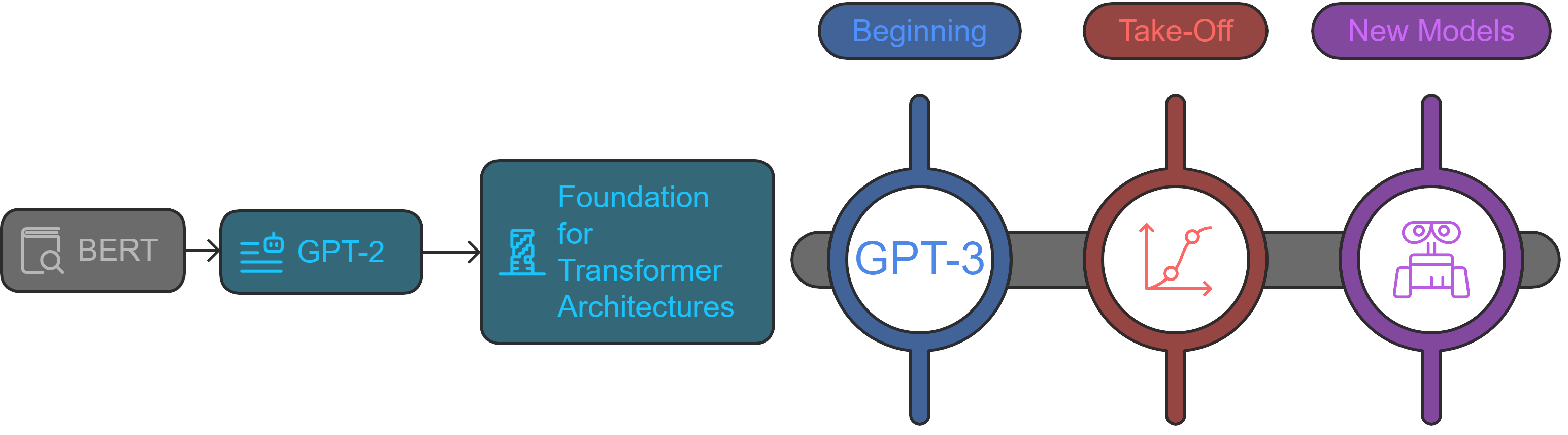}
    \caption{LLMs Beginning and Evolution}
    \label{fig:beginning_and_evol}
\end{figure}

The rapid progression within model families, such as OpenAI's GPT series (from GPT-3.5 to GPT-4 and beyond), Google's Gemini models, Anthropic's Claude series, and Meta's LLaMA versions, has resulted in substantial improvements in accuracy, contextual understanding, and reasoning capabilities. This is notably demonstrated by the performance of models like GPT-4 on medical licensing assessments \cite{b4}. For instance, on previous editions of the Portuguese PNA (2019-2023), ChatGPT-4o significantly outperformed its predecessor, ChatGPT-3.5, achieving scores sufficient for any medical specialty placement \cite{b2}, foreshadowing the capabilities tested in our current study with even newer models. Figure \ref{fig:beginning_and_evol} is a short diagram that represents this evolution. 

\subsection{The Current LLM Landscape}
\label{subsec:landscape}
The field is rapidly evolving, with numerous organizations releasing powerful models. A comparison across various LLMs reveals a wide spectrum of capabilities, architectural differences (e.g., dense vs. Mixture-of-Experts), and access models (open-weight vs. API-only). Understanding the characteristics of these models, such as their knowledge cutoff dates, context window sizes, and licensing terms (as detailed in Table \ref{tab:llm_metadata}), is crucial for evaluating their suitability for specific healthcare applications. Detailed metadata for each evaluated LLM, grouped by developer, is provided in Table \ref{tab:llm_metadata} on the following landscape page.

\section{Reasoning Methodologies Enhancements}
\label{sec:reasoning}
Significant advancements in reasoning methodologies are reshaping how LLMs perform complex tasks, including those relevant to diagnostics and therapeutic planning.

\subsection{Chain of Thought (CoT)}
\begin{figure}[h!]
    \centering
    \includegraphics[width=0.5\linewidth]{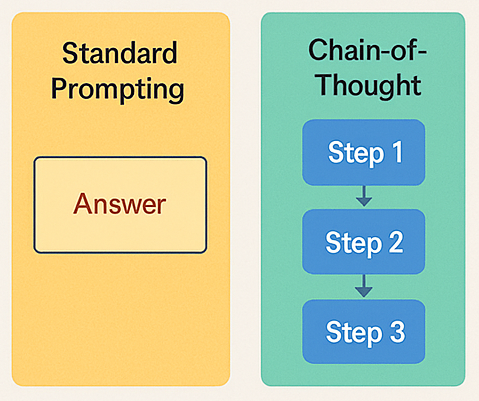}
    \caption{Chain of Thought}
    \label{fig:Chain of Thought}
\end{figure}

The Chain of Thought (CoT) approach encourages LLMs to generate intermediate reasoning steps before arriving at a final answer, mimicking human cognitive processes. This has been shown to significantly improve performance in logical, mathematical, and complex reasoning tasks \cite{yang2023gptsolvemathematicalproblems}, making it particularly relevant for structured clinical diagnosis scenarios and multi-stage medical reasoning challenges. Several models evaluated in this study incorporate explicit CoT or "thinking" modes (indicated by $^\star$ in Table \ref{tab:llm_performance}). Figure \ref{fig:Chain of Thought} represents this mechanism. 

\newpage
\subsection{Chain of Draft (CoD)}
While Chain of Thought (CoT) significantly improves reasoning by encouraging verbose, step-by-step explanations, this approach often leads to high token usage and increased latency, contrasting with efficient human problem-solving \cite{b5}. Humans typically rely on concise intermediate notes or drafts, capturing only essential information to guide their thinking process.

Inspired by this cognitive efficiency, Xu et al. proposed \textbf{Chain of Draft (CoD)} \cite{b5}. This paradigm prompts LLMs to generate \textit{minimalistic yet informative} intermediate reasoning outputs—essentially concise "drafts"—at each step, rather than fully elaborated explanations. The goal is to focus computational effort on critical insights and calculations necessary to advance towards the solution, stripping away redundant verbosity.

Experiments across arithmetic, commonsense, and symbolic reasoning tasks demonstrated that CoD can match or even surpass the accuracy of standard CoT prompting \cite{b5}. Crucially, this performance parity is achieved with a dramatic reduction in resource consumption, using significantly fewer tokens (reportedly as little as 7.6\% of CoT tokens in some cases) and thereby lowering both inference cost and latency. This efficiency makes CoD a potentially valuable strategy for deploying LLMs in resource-constrained or time-sensitive environments, such as real-time clinical decision support, where rapid and accurate reasoning is paramount but computational overhead must be minimized. However, the authors note that CoD's effectiveness, particularly in zero-shot scenarios or with smaller models, can depend on the model's underlying training and may benefit from few-shot examples or specific fine-tuning \cite{b5}.


\subsection{Emerging Diffusion-Based Language Models}
\label{subsec:diffusion}
While the models evaluated in this study predominantly use autoregressive architectures (generating text sequentially token by token), a newer generation of language models based on \textbf{diffusion processes} is emerging. Inspired by the success of diffusion models in image and video generation (e.g., DALL-E, Midjourney, Sora) \cite{diffusion_survey_ref}, this approach, fundamentally different from sequential generation, is being adapted for text tasks \cite{b6}.

Companies like Inception Labs are pioneering commercial-scale diffusion Large Language Models (dLLMs), such as their announced Mercury model \cite{inception_mercury_ref}, claiming significant potential advantages over traditional autoregressive LLMs \cite{inception_website_ref}. Proposed benefits often highlighted include:
\begin{itemize}
    \item \textbf{Speed and Efficiency:} Diffusion processes can allow for parallel generation steps, potentially offering substantially faster inference (with claims of 5-10x improvements) and reduced computational cost compared to the inherently sequential nature of autoregressive models.
    \item \textbf{Reasoning and Reliability:} It is suggested that some diffusion frameworks might possess built-in mechanisms for iterative refinement or error correction during the generation process. This could potentially lead to improved reasoning capabilities and a reduction in common LLM issues like hallucinations, although this requires rigorous validation.
    \item \textbf{Output Control:} The generative process in diffusion models may afford enhanced control over the structure, style, or specific constraints of the generated text, making them potentially well-suited for tasks requiring precise formats, such as function calling or structured data generation (e.g., filling medical templates).
    \item \textbf{Multimodality:} The underlying mathematical framework of diffusion models might offer a more unified approach for generative AI across different data types, including text, images, audio, and video, potentially simplifying the development of truly multimodal systems.
\end{itemize}
Although dLLMs were not included in our current evaluation due to their novelty and limited availability/benchmarking at the time of testing, they represent a potentially significant future direction in LLM development. Their distinct generative process could lead to different performance characteristics and trade-offs compared to current models. Further research and independent benchmarking, especially on complex medical reasoning and generation tasks, are required to validate these claims and understand their applicability and safety within clinical contexts. Their development certainly warrants close attention as the field progresses towards potentially faster, more controllable, and perhaps more robust language generation systems.

\subsection{Insights from Leading AI Experts}
Analyzing opinions from key AI leaders contextualizes broader issues relevant to LLM deployment in sensitive areas like healthcare more precisely.

\subsubsection{Andrej Karpathy’s Concept of “Jagged Intelligence”}
Karpathy coined “Jagged Intelligence,” highlighting the inconsistency of model performance, even among state-of-the-art LLMs. Models often excel at complex scenarios but fail at simpler logical checks, an issue attributable to training distribution biases \cite{karpathy_jagged_intelligence_ref}. Such inconsistencies pose significant problems for healthcare applications requiring dependable accuracy across varied scenarios.

\subsubsection{Ilya Sutskever’s Emphasis on Reinforcement Learning}
OpenAI’s Ilya Sutskever underscores reinforcement learning (RL) approaches \cite{lightman2023letsverifystepstep}, such as Prover-Verifier Games (PVG) \cite{Kirchner2024ProverVerifierGames}, for developing more consistent, truthful, and ethically sound LLMs. Integrating RL into healthcare-oriented LLMs presents a viable path to robust medical AI.

\subsubsection{Yann LeCun on AI-Human Symbiosis}
LeCun argues for AI systems augmenting rather than replacing human professionals, ideally supporting clinical judgment, thus safeguarding ethical and professional standards \cite{lecun_fridman_2023}. This aligns with the goal of using LLMs as assistive tools in diagnosis and treatment.

\subsubsection{Demis Hassabis on AI-Driven Medical Innovations}
DeepMind CEO Demis Hassabis views AI advances through a lens of collaborative potential. Successes like AlphaFold \cite{alphafold2021} demonstrate that innovations originating from AI can profoundly accelerate medical breakthroughs and transform patient care, suggesting LLMs could plays a similar transformative role in clinical decision support. This innovation's significance was recognized with the 2024 Nobel Prize in Chemistry, awarded to David Baker, Demis Hassabis, and John Jumper \cite{ch_2025_nobel_prize}.

\section{Challenges and Ethical Considerations in Clinical Deployment}
\label{sec:challenges}
Despite technological advancements, the implementation of LLMs as diagnostic and treatment aids in clinical settings faces several fundamental challenges:

\begin{itemize}
    \item \textbf{Accuracy and Reliability:} Ensuring the model's outputs are consistently accurate, reliable, and robust across diverse clinical scenarios and patient populations. Addressing the "jagged intelligence" problem \cite{karpathy_jagged_intelligence_ref} is critical.
    \item \textbf{Data Privacy and Security:} Protecting sensitive patient health information (PHI) is paramount, especially when using cloud-based APIs or models trained on broad datasets. Compliance with regulations like HIPAA (in the US) or GDPR (in the EU) is essential.
    \item \textbf{Clinical Workflow Integration:} Introducing LLM-based tools seamlessly into existing clinical workflows and electronic health record (EHR) systems without disrupting care or increasing clinician burden.
    \item \textbf{Validation and Regulation:} Rigorously validating model performance in real-world clinical settings and navigating the regulatory landscape for AI-driven medical devices or clinical decision support software. In the European Union, compliance with emerging regulations like the \textbf{EU AI Act}—which may classify certain AI-based medical applications as high-risk—will be essential for clinical deployment.
    \item \textbf{Bias and Equity:} Identifying and mitigating potential biases in LLMs (stemming from training data) that could lead to health disparities or inequitable care.
    \item \textbf{Transparency and Explainability:} Providing clinicians with understandable justifications for AI-generated suggestions (especially when using complex models or CoT/CoD reasoning) to enable informed clinical judgment.
    \item \textbf{Ethics and Accountability:} Defining clear ethical guidelines for the use of LLMs in patient care, establishing lines of accountability when errors occur, and ensuring patient consent and understanding.
\end{itemize}

Addressing these challenges requires a multidisciplinary approach involving AI researchers, clinicians, ethicists, regulators, and patients.
\section{Methodology and Results}
\label{sec:results}

\subsection{Evaluation Benchmark and Setup}
\label{subsec:setup}
In this study, tests were carried out to assess the performance of various LLMs on a standardized medical knowledge assessment. The 2024 Portuguese National Exam for Access to Specialty (\textit{Prova Nacional de Acesso} - PNA) \cite{pna2024}, containing 150 multiple-choice questions covering diverse medical fields, was used as the benchmark. This exam evaluates a broad range of medical knowledge expected for specialty access in Portugal. It is important to note that these models were not specifically fine-tuned on this exam dataset, providing a measure of their generalized medical knowledge and reasoning capabilities, particularly in the Portuguese language context.

\begin{figure}
    \centering
    \includegraphics[width=0.3\linewidth]{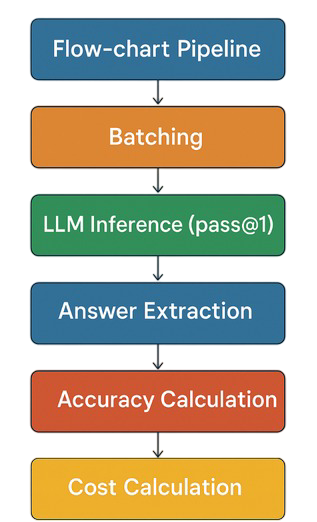}
    \caption{Methodology Flowchart}
    \label{fig:enter-label}
\end{figure}

The LLMs were evaluated using a strict \textbf{pass@1 methodology}, where only the first generated answer for each question was evaluated, with no subsequent attempts or retries permitted. This approach ensures a conservative performance assessment, reflective of scenarios where an initial output might be acted upon or considered definitive, demanding high reliability from the first response. The questions were presented to the models in batches of 10. Testing was constrained by the maximum token limit supported by each model's context window; for models with context windows exceeding 100,000 tokens, the effective limit for this test was capped around this value to ensure comparability and manage computational resources. Figure \ref{fig:enter-label} shows the methodology flowchart. 

For all evaluations, no specialized prompt engineering or model-specific instructions were used. Each model was presented with the exam questions exactly as written, split into batches of 10 questions each (15 batches in total for the 150-question exam). A Python script automated the process of submitting questions in batches, collecting model outputs, and evaluating the answers. No few-shot examples, explicit reasoning prompts, or advanced formatting were provided; models were simply asked to respond directly to each question. Consequently, the results reported here are conservative estimates of model performance. It is likely that accuracy could be further improved - potentially by several percentage points (\(\sim\){[2\% - 5\%]}) - by applying prompt engineering methods such as few-shot examples, Chain-of-Thought instructions, or other optimization strategies. In contrast to prior studies which leverage advanced prompt engineering techniques, our approach used no additional instructions or context; models received only the original question text in each prompt.

\newpage
\subsection{Performance Scoring Metric}
\label{subsec:scoring}

To provide a holistic evaluation of the Large Language Models (LLMs), we developed a custom scoring metric. This metric goes beyond simple prediction accuracy to incorporate the crucial factor of cost-effectiveness, which is vital for real-world deployment viability. Furthermore, it includes a minor adjustment for potential data contamination risk.

The final composite score is calculated using the following formula:

\begin{equation}
\label{eq:score_final}
\text{Score} = 100 \times \left(\frac{\text{Correct}}{N}\right)^3 \times \frac{1}{\sqrt{1 + \log_{10}(P + 1)}} \times C_{\text{risk}}
\end{equation}

Where the components are defined as:
\begin{itemize}
    \item \texttt{Correct}: The number of questions the model answered correctly.
    \item \texttt{N}: The total number of questions in the evaluation set (here, $N = 150$).
    \item \texttt{P}: The approximate price in USD per 1 million tokens for using the model. This typically uses a blend of input/output pricing relevant for balanced tasks, based on publicly available data around Q1 2025. For models that are free, open-source, or lack clear public pricing (e.g., experimental versions) at the time of testing, we used $P=0$. (Specific prices used are detailed in Table~\ref{tab:llm_performance}). 
    \item \texttt{C\_risk}: A contamination risk penalty factor, applied to account for the possibility that the model might have inadvertently "seen" or memorized parts of the evaluation data during its training, especially if its knowledge cut-off date is after the exam's creation date. The penalty is applied based on an assessed risk level:
    \begin{itemize}
        \item \texttt{1.0} for Low risk (\contamLow)
        \item \texttt{0.95} for Medium risk (\contamMed)
        \item \texttt{0.90} for High risk (\contamHigh)
    \end{itemize}
    This penalty is deliberately gentle, acknowledging the difficulty in definitively proving contamination and the possibility of genuine knowledge acquisition.
\end{itemize}

\subsubsection{\textbf{Explanation of Formula Components:}}

The formula is designed to balance three key aspects:

\begin{enumerate}
    \item \textbf{Accuracy Component:} $\left(\frac{\text{Correct}}{N}\right)^3$
    \begin{itemize}
        \item This term represents the model's raw accuracy ($\textit{Correct}/\textit{N}$), raised to the power of 3.
        \item \textit{Rationale:} Cubing the accuracy ratio (a value between 0 and 1) heavily emphasizes high performance. It means that improvements in accuracy at the higher end (e.g., going from 80\% to 90\%) result in a much larger increase in the score compared to gains at the lower end (e.g., going from 30\% to 40\%). This design choice strongly rewards models achieving top-tier correctness, reflecting the high stakes often associated with clinical applications.
    \end{itemize}

    \item \textbf{Cost-Efficiency Component:} $\frac{1}{\sqrt{1 + \log_{10}(P + 1)}}$
    \begin{itemize}
        \item This term acts as a penalty factor based on the model's price \texttt{P}.
        \item \textit{Rationale:}
        \begin{itemize}
            \item $\log_{10}(P + 1)$: The logarithm is used to compress the wide range of potential model prices (from \$0 to over \$100 per million tokens). Adding 1 (i.e., $P + 1$) ensures that the logarithm is always defined, even for free models where $P=0$ (since $\log_{10}(1) = 0$). This prevents the score from becoming undefined or infinitely large for free models.
            \item $\sqrt{\dots}$: Taking the square root of the logarithmic term moderates the cost penalty. It acknowledges that higher costs might be justified if they deliver substantially better accuracy, but still ensures that more economical models are favoured, all else being equal. The entire term acts as a multiplier between 0 and 1 (it is 1 when $P=0$, and decreases as $P$ increases).
        \end{itemize}
    \end{itemize}

    \item \textbf{Contamination Risk Adjustment:} $C_{\text{risk}}$
    \begin{itemize}
        \item This is a simple multiplier that slightly reduces the score for models assessed to have a medium or high risk of data contamination.
        \item \textit{Rationale:} It serves as a small corrective factor to temper the scores of models that might have had an unfair advantage due to exposure to the evaluation data. The mildness of the penalty (maximum 10\% reduction) reflects the uncertainty inherent in contamination assessment.
    \end{itemize}
\end{enumerate}
\subsubsection{\textbf{Interpretation:}}

The final score, theoretically ranging from 0 to 100 (though the maximum achieved in this study was approximately \MaxScore), provides a single, composite metric reflecting the overall practical utility of each LLM for the specific task. A higher score indicates a more desirable balance between high accuracy, reasonable cost, and low contamination risk, making it a better candidate for potential real-world, cost-sensitive applications like those in clinical settings.
\subsection{LLM Performance Results}
\label{subsec:pna_results}
The performance of each evaluated LLM on the PNA 2024 exam, along with its associated cost and the calculated overall score (Eq. \ref{eq:score_final}), is presented in Figure \ref{fig:llm_performance_chart_appendix} and detailed in Table \ref{tab:llm_performance}. For context, the performance of the highest-scoring human candidate and the average human candidate on the same exam are included as reference points in the table.



\begin{table*}[htbp]
\centering
\small
\renewcommand{\arraystretch}{1.1}
\caption{\textbf{LLM Performance on PNA 2024 Exam (150 Questions) vs. Human Benchmarks}}
\label{tab:llm_performance}
\begin{tabular}{
  >{\raggedright\arraybackslash}p{3.8cm} 
  S[table-format=3.0] 
  S[table-format=2.0] 
  >{\raggedleft\arraybackslash}p{1.8cm} 
  S[table-format=2.2] 
  S[table-format=2.5] 
  >{\centering\arraybackslash}p{1.5cm} 
  >{\centering\arraybackslash}p{3.2cm} 
}
\toprule
\textbf{Model} &
{\textbf{Correct}} &
{\textbf{Wrong}} &
\textbf{Price (\$ / 1M tok)} &
{\textbf{Score}} &
{\textbf{Score\_Final}} &
{\textbf{Contam. Risk}} &
\textbf{Performance (\% of Max Score)} \\

\midrule
\multicolumn{8}{l}{\textbf{Google}} \\
\rowcolor{abovebench} Gemini 2.5 Pro (Exp)$^{\star}$ & 135 & 15 & \$0.00 & 72.90 & 65.61 & \contamHigh & 
\begin{tikzpicture}[baseline=0pt]\fill[perfbest] (0,0) rectangle ({3 * 65.61 / 65.61}, 0.24);\end{tikzpicture} \\
\rowcolor{abovebench} Gemini 2.0 Flash Thinking (Exp)$^{\star}$ & 128 & 22 & \$0.00 & 62.14 & 62.14 & \contamLow & 
\begin{tikzpicture}[baseline=0pt]\fill[black!10] (0,0) rectangle (3,0.24);\fill[black!80] (0,0) rectangle ({3 * 62.14 / 65.61}, 0.24);\end{tikzpicture} \\
\rowcolor{abovebench} Gemini 2.0 Pro (Exp 02-05) & 125 & 25 & \$0.00 & 57.87 & 52.08 & \contamHigh & 
\begin{tikzpicture}[baseline=0pt]\fill[black!10] (0,0) rectangle (3,0.24);\fill[black!85] (0,0) rectangle ({3 * 52.08 / 65.61}, 0.24);\end{tikzpicture} \\
\rowcolor{belowbench} Gemma 3 27B & 94 & 56 & \$0.00 & 24.61 & 24.61 & \contamLow & 
\begin{tikzpicture}[baseline=0pt]\fill[black!10] (0,0) rectangle (3,0.24);\fill[perfmostwrong] (0,0) rectangle ({3 * 24.61 / 65.61}, 0.24);\end{tikzpicture} \\
\midrule
\multicolumn{8}{l}{\textbf{Meta}} \\
\rowcolor{highlightrow} LLaMA 4 Maverick & 120 & 30 & \$0.00 & 51.20 & 51.20 & \contamLow & 
\begin{tikzpicture}[baseline=0pt]\fill[black!10] (0,0) rectangle (3,0.24);\fill[black!85] (0,0) rectangle ({3 * 51.20 / 65.61}, 0.24);\end{tikzpicture} \\
\rowcolor{highlightrow} LLaMA 3.1 405B & 112 & 38 & \$0.00 & 41.63 & 41.63 & \contamLow & 
\begin{tikzpicture}[baseline=0pt]\fill[black!10] (0,0) rectangle (3,0.24);\fill[black!85] (0,0) rectangle ({3 * 41.63 / 65.61}, 0.24);\end{tikzpicture} \\
\rowcolor{highlightrow} LLaMA 3.3 70B (Instruct) & 109 & 41 & \$0.00 & 38.37 & 38.37 & \contamLow & 
\begin{tikzpicture}[baseline=0pt]\fill[black!10] (0,0) rectangle (3,0.24);\fill[black!85] (0,0) rectangle ({3 * 38.37 / 65.61}, 0.24);\end{tikzpicture} \\

\midrule
\multicolumn{8}{l}{\textbf{Deepseek}} \\
\rowcolor{highlightrow} DeepSeek R1$^\star$ & 121 & 29 & \$0.00 & 52.49 & 47.24 & \contamHigh & 
\begin{tikzpicture}[baseline=0pt]\fill[black!10] (0,0) rectangle (3,0.24);\fill[black!85] (0,0) rectangle ({3 * 47.24 / 65.61}, 0.24);\end{tikzpicture} \\
\rowcolor{highlightrow} DeepSeek V3 & 110 & 40 & \$0.00 & 39.44 & 37.47 & \contamMed & 
\begin{tikzpicture}[baseline=0pt]\fill[black!10] (0,0) rectangle (3,0.24);\fill[black!85] (0,0) rectangle ({3 * 37.47 / 65.61}, 0.24);\end{tikzpicture} \\
\midrule

\multicolumn{8}{l}{\textbf{OpenAI}} \\
\rowcolor{abovebench} O3-Mini (High mode)$^\star$ & 126 & 24 & \$3.74 & 45.79 & 45.79 & \contamLow & 
\begin{tikzpicture}[baseline=0pt]\fill[black!10] (0,0) rectangle (3,0.24);\fill[black!85] (0,0) rectangle ({3 * 45.79 / 65.61}, 0.24);\end{tikzpicture} \\
\rowcolor{exceedsmax} O1$^{\star}$ & 136 & 14 & \$51.00 & 45.22 & 45.22 & \contamLow & 
\begin{tikzpicture}[baseline=0pt]\fill[black!10] (0,0) rectangle (3,0.24);\fill[perfmostcorrect] (0,0) rectangle ({3 * 45.22 / 65.61}, 0.24);\end{tikzpicture} \\
\rowcolor{abovebench} GPT-4.5 (Preview) & 133 & 17 & \$135.00 & 39.38 & 39.38 & \contamLow & 
\begin{tikzpicture}[baseline=0pt]\fill[black!10] (0,0) rectangle (3,0.24);\fill[black!85] (0,0) rectangle ({3 * 39.38 / 65.61}, 0.24);\end{tikzpicture} \\
\rowcolor{abovebench} ChatGPT GPT-4o (Latest) & 124 & 26 & \$8.50 & 40.17 & 38.16 & \contamMed & 
\begin{tikzpicture}[baseline=0pt]\fill[black!10] (0,0) rectangle (3,0.24);\fill[black!85] (0,0) rectangle ({3 * 38.16 / 65.61}, 0.24);\end{tikzpicture} \\
\rowcolor{belowbench} GPT-4o Mini & 99 & 51 & \$0.51 & 26.48 & 26.48 & \contamLow & 
\begin{tikzpicture}[baseline=0pt]\fill[black!10] (0,0) rectangle (3,0.24);\fill[black!85] (0,0) rectangle ({3 * 26.48 / 65.61}, 0.24);\end{tikzpicture} \\

\midrule
\multicolumn{8}{l}{\textbf{Anthropic}} \\
\rowcolor{abovebench} Claude 3.7 Sonnet (Thinking)$^{\star}$ & 129 & 21 & \$18.00 & 42.14 & 40.03 & \contamMed & 
\begin{tikzpicture}[baseline=0pt]\fill[black!10] (0,0) rectangle (3,0.24);\fill[black!85] (0,0) rectangle ({3 * 40.03 / 65.61}, 0.24);\end{tikzpicture} \\
\rowcolor{abovebench} Claude 3.5 Sonnet (New) & 125 & 25 & \$12.60 & 39.62 & 39.62 & \contamLow & 
\begin{tikzpicture}[baseline=0pt]\fill[black!10] (0,0) rectangle (3,0.24);\fill[black!85] (0,0) rectangle ({3 * 39.62 / 65.61}, 0.24);\end{tikzpicture} \\
\rowcolor{highlightrow} Claude 3.5 Haiku & 107 & 43 & \$3.36 & 28.35 & 28.35 & \contamLow & 
\begin{tikzpicture}[baseline=0pt]\fill[black!10] (0,0) rectangle (3,0.24);\fill[black!85] (0,0) rectangle ({3 * 28.35 / 65.61}, 0.24);\end{tikzpicture} \\

\midrule
\multicolumn{8}{l}{\textbf{XAI}} \\
\rowcolor{abovebench} Grok 3 (Thinking Mode)$^{\star}$ & 126 & 24 & \$5.00 & 44.45 & 42.23 & \contamMed & 
\begin{tikzpicture}[baseline=0pt]\fill[black!10] (0,0) rectangle (3,0.24);\fill[black!85] (0,0) rectangle ({3 * 42.23 / 65.61}, 0.24);\end{tikzpicture} \\

\midrule
\multicolumn{8}{l}{\textbf{Others (Alibaba, Mistral)}} \\
\rowcolor{highlightrow} Alibaba Qwen 2.5-Max$^{\star}$ & 117 & 33 & \$5.40 & 35.31 & 35.31 & \contamLow & 
\begin{tikzpicture}[baseline=0pt]\fill[black!10] (0,0) rectangle (3,0.24);\fill[black!85] (0,0) rectangle ({3 * 35.31 / 65.61}, 0.24);\end{tikzpicture} \\
\rowcolor{highlightrow} Mistral Small & 102 & 48 & \$0.26 & 29.97 & 29.97 & \contamLow & 
\begin{tikzpicture}[baseline=0pt]\fill[black!10] (0,0) rectangle (3,0.24);\fill[black!85] (0,0) rectangle ({3 * 29.97 / 65.61}, 0.24);\end{tikzpicture} \\
\rowcolor{highlightrow} Mistral Large 24.11 & 110 & 40 & \$5.20 & 29.46 & 27.99 & \contamMed & 
\begin{tikzpicture}[baseline=0pt]\fill[black!10] (0,0) rectangle (3,0.24);\fill[black!85] (0,0) rectangle ({3 * 27.99 / 65.61}, 0.24);\end{tikzpicture} \\

\midrule
\multicolumn{8}{l}{\textbf{Reference Points}} \\
\multicolumn{1}{l}{Top Student (Max Score)} & 135 & 15 & {—} & {—} & {N/A} & {N/A} &
\begin{tikzpicture}[baseline=0pt] \fill[lightgray] (0,0) rectangle (3,0.24); \draw[dashed, thick, red] (0,0.12) -- ({3*135/150},0.12); \node[right, red, font=\tiny] at ({3*135/150+0.1}, 0.12) {Max (135)}; \end{tikzpicture} \\
\multicolumn{1}{l}{95th Percentile Student} & 122 & 28 & {—} & {—} & {N/A} & {N/A} &
\begin{tikzpicture}[baseline=0pt] \fill[lightgray] (0,0) rectangle (3,0.24); \draw[dotted, thick, orange] (0,0.12) -- ({3*122/150},0.12); \node[right, orange, font=\tiny] at ({3*122/150+0.1}, 0.12) {95th Pctl (122)}; \end{tikzpicture} \\
\multicolumn{1}{l}{Median Student (50th Pctl)} & 101 & 49 & {—} & {—} & {N/A} & {N/A} &
\begin{tikzpicture}[baseline=0pt] \fill[lightgray] (0,0) rectangle (3,0.24); \draw[dotted, thick, blue] (0,0.12) -- ({3*101/150},0.12); \node[right, blue, font=\tiny] at ({3*101/150+0.1}, 0.12) {Median (101)}; \end{tikzpicture} \\
\multicolumn{1}{l}{LLM Average (This Study)} & 118 & 32 & \$11.95 & 41.30 &    40.43 & {Mixed} & \\
\bottomrule
\end{tabular}
\vspace{0.2cm}
\begin{minipage}{\textwidth}
\small
\vspace{10pt}
\textbf{Notes:}
\begin{itemize}
    \item[$^\star$] Indicates models tested using an explicit Chain-of-Thought (CoT) / "Thinking" mode or models known for strong internal reasoning. Performance may differ in standard modes.
    \item Colors indicate LLM performance relative to human PNA 2024 student benchmarks (Correct Answers):
        \colorbox{exceedsmax}{Exceeds Max Student (>135 Correct)},
        \colorbox{abovebench}{95th Pctl to Max Student (122-135 Correct)},
        \colorbox{highlightrow}{Median to 95th Pctl Student (101-121 Correct)},
        \colorbox{belowbench}{Below Median Student (<101 Correct)}.
   \item Price is estimated cost in USD per 1 million tokens, calculated assuming a typical workload mix (e.g., weighted as 80\% output tokens and 20\% input tokens, $P = 0.8 \times P_{output} + 0.2 \times P_{input}$ per 1M tokens), based on public pricing data around Q1 2025. Pricing structures vary by provider. Free/open-source models shown as \$0.00.
    \item \textbf{Potential Contamination Risk:} (\contamLow{} Low, \contamMed{} Med, \contamHigh{} High). See Sec. \ref{subsec:cutoff_contamination}.
     \item \textbf{Performance bars:} \raisebox{0pt}{\tikz\draw[perfbest,fill=perfbest] (0,0) rectangle (.6,0.2);} Best score;
        \raisebox{0pt}{\tikz\draw[perfmostcorrect,fill=perfmostcorrect] (0,0) rectangle (.6,0.2);} Most correct answers;
        \raisebox{0pt}{\tikz\draw[perfmostwrong,fill=perfmostwrong] (0,0) rectangle (.6,0.2);} Most incorrect answers. Scaled to max LLM score (65.61).
    \item See Figure \ref{fig:me} for a visual comparison of top LLM scores against the student performance distribution.
\end{itemize}
\end{minipage}
\end{table*}


\begin{sidebox}[green!8]{\textbf{At-a-glance: Human vs LLM on PNA 2024}}
\small
\textbf{Top Student (Max Score):} 135/150 $(90\%)$ \\ 
\textbf{Median Student (50th Pctl):} 101/150 $(67\%)$ \\ 
\textbf{95th Percentile Student:} 122/150 $(81\%)$ \\ 
\textbf{Best LLM (Correct Answers):} 136/150 $(91\%)$ \\ 
\\
$\rightarrow$ Many LLMs exceed the median student score (101). Several LLMs outperform the 95th percentile student (122), with the top LLM slightly surpassing the maximum student score. 
\end{sidebox}
\newpage
\section{Discussion}
\label{sec:discussion}
\subsection{Impact of Knowledge Cutoff Dates and Potential Data Contamination}
\label{subsec:cutoff_contamination}
A significant factor influencing LLM performance on benchmarks is the model's \textbf{knowledge cutoff date} (see Table \ref{tab:llm_metadata}), representing the point up to which its training data extends. Our benchmark, the PNA 2024 exam, was administered in \textbf{November 2024}. Exam questions and answers likely became available online sometime \textit{after} this date.

Consequently, models with knowledge cutoff dates extending significantly beyond November 2024 theoretically \textit{could} have encountered the PNA 2024 materials if these were scraped from the public web and included in subsequent training runs. If such data contamination occurred, the model's performance on the exam might reflect memorization rather than genuine medical knowledge application or reasoning ability.

While major AI developers often employ filtering techniques to prevent contamination from known benchmark datasets, the recency of the exam relative to the cutoff dates of the latest models makes absolute exclusion difficult to guarantee without transparency into training datasets and filtering protocols.

To acknowledge this potential issue, we have added a "Contam. Risk" column in Table \ref{tab:llm_performance}. This column uses a scale of colored dots (\contamLow{}, \contamMed{}, \contamHigh{}) to visually indicate the \textit{theoretical possibility} of contamination based on the temporal overlap between the model's knowledge cutoff and the benchmark's likely online availability (post-November 2024):
\begin{itemize}
    \item[\contamLow{}] \textbf{Low Risk} (Green Dot): Indicates the model's knowledge cutoff is earlier than November 2024. Contamination with PNA 2024 materials is highly unlikely as they were not yet available online during the training data collection period.
    \item[\contamMed{}] \textbf{Medium Risk} (Amber Dots): Indicates the cutoff is shortly after November 2024 (e.g., Nov 2024, Dec 2024). Contamination is possible if exam data appeared online very quickly and was included in training datasets collected during this narrow window.
    \item[\contamHigh{}] \textbf{High Risk} (Red Dots): Indicates the cutoff is significantly later (e.g., Jan 2025 or later), providing a longer timeframe during which the PNA 2024 materials could potentially have been available online and ingested into training data.
\end{itemize}
It is crucial to interpret this column simply as an indicator of temporal overlap and theoretical risk based on publicly available information. It does \textbf{not} confirm data contamination. Nonetheless, this temporal context is important when evaluating the reported performance figures, especially for models rated Medium (\contamMed{}) or High (\contamHigh{}) risk, as their scores might be partially inflated if contamination occurred. Models rated Low (\contamLow{}) risk provide a more reliable assessment of reasoning on unseen data in this context.
\subsection{Performance-Cost Relationship for Clinical Implementation}
Our results, visualized in Figure \ref{fig:llm_performance_chart_appendix} and detailed in Table \ref{tab:llm_performance}, reveal a complex relationship between LLM performance and cost in the context of medical knowledge assessment. Notably, some of the highest overall scores, which balance accuracy and cost via Eq. \ref{eq:score_final}, were achieved by models available at low or no cost. Google's experimental Gemini 2.5 Pro (Score: \MaxScore, assuming negligible preview cost) and Gemini 2.0 Flash Thinking (Score: 62.14, \$0.00) lead the rankings, alongside the open-source DeepSeek R1 (Score: 52.49, \$0.00). This suggests that powerful AI assistance for tasks like preliminary diagnostic support or information retrieval might be accessible even within constrained healthcare budgets.

Conversely, while some high-cost models like OpenAI's O1 (raw accuracy leader with 136 correct answers, 90.7\%) and GPT-4.5 Preview (133 correct, 88.7\%) achieved top-tier accuracy, their substantial price (\$51.00 and an estimated \$135.00 per 1M tokens, respectively) significantly lowered their overall utility score in our cost-sensitive evaluation (Scores: 45.22 and 39.38). This highlights a critical consideration for widespread clinical adoption: the need for solutions that are not only accurate but also economically sustainable.

Our scoring methodology deliberately favors cost-efficiency, bringing models like OpenAI's O3-Mini-High (Score: 45.79, \$3.74) and Grok 3 Thinking (Score: 44.45, \$12.60) into prominence as strong performers offering a good balance between capability and cost.

\subsection{Impact of Reasoning Methodologies}
The evaluation indicates a potential advantage for models employing explicit reasoning steps (marked with $^\star$ in Table \ref{tab:llm_performance}). Several of the top-performing models, including the leading Google Gemini models, DeepSeek R1, O3-Mini-High, Claude 3.7 Sonnet Thinking, and Grok 3 Thinking, utilize Chain-of-Thought or similar internal "thinking" processes. This suggests that structured reasoning may contribute to higher accuracy on complex, multi-step problems like those found in medical exams. Techniques like CoD \cite{b5}, aiming for similar accuracy with greater efficiency, further highlight the importance of structured yet potentially concise reasoning.

For clinical applications, the transparency potentially offered by CoT/CoD outputs could be invaluable. Clinicians need to understand the rationale behind an AI's suggestion to trust and appropriately utilize it. Models that can "show their work" may integrate better into clinical decision-making workflows, allowing for verification and building confidence in the AI as a supportive tool. However, the potential latency and token costs associated with verbose reasoning need to be balanced against the speed and efficiency requirements of clinical settings.

\subsection{Reliability and Hallucinations}
A critical aspect not fully captured by standardized test accuracy is the propensity of LLMs to "hallucinate" – generating plausible but factually incorrect or nonsensical information. While models like GPT-4 and its successors have improved in factual grounding compared to earlier iterations, the risk remains, particularly in high-stakes domains like medicine \cite{b3}. Confidently incorrect outputs, misinterpretation of nuanced clinical data, or failures in complex logical chains (even amidst overall high performance, i.e., "jagged intelligence" \cite{karpathy_jagged_intelligence_ref}) can have serious consequences. Our pass@1 methodology partially reflects the need for initial reliability, but it doesn't measure the frequency or nature of potential errors. This underscores the non-negotiable requirement for vigilant human oversight; LLMs should function as assistants providing suggestions and information for review, not as autonomous decision-makers.

\begin{sidebox}[red!7]{Common LLM Mistakes ("Hallucinations")}
\small
Example: LLM recommends an obsolete/off-market drug for tuberculosis (TB), or fails an easy matching question, showing ``jagged intelligence.'' Both plausible sounding, factually incorrect.
\end{sidebox}

\subsection{Provider Landscape and Model Characteristics}
Distinct patterns emerge across different AI providers, linking performance to model characteristics detailed in Table \ref{tab:llm_metadata}:
\begin{enumerate}
    \item \textbf{Google:} Showed exceptional performance with its latest Gemini models, particularly the experimental versions featuring large context windows (up to 1M tokens) and recent knowledge cutoffs (Table \ref{tab:llm_metadata}), achieving high accuracy at potentially very low costs.
    \item \textbf{OpenAI:} Fields highly accurate models (O1, GPT-4.5, GPT-4o), often setting the benchmark for raw capability, but premium pricing impacts cost-effectiveness scores. O3-Mini offers a competitive mid-tier option.
    \item \textbf{Anthropic:} Models demonstrate strong performance, particularly with explicit reasoning modes (Claude 3.7 Sonnet Thinking), often balancing accuracy with safety considerations.
    \item \textbf{DeepSeek:} Impressed with DeepSeek R1, a high-performing open-source model leveraging a large Mixture-of-Experts (MoE) architecture (Table \ref{tab:llm_metadata}), highlighting the increasing capability of non-commercial alternatives.
    \item \textbf{Meta:} The LLaMA models showed solid performance as source-available options, though they were outperformed by several competitors on this specific task and scoring metric.
    \item \textbf{Others:} Models from Alibaba (Qwen), Mistral (Large, Small), and xAI (Grok) also demonstrated competence, contributing to a diverse landscape with varying strengths (e.g., Mistral's efficiency, Grok's real-time data integration).
\end{enumerate}
Factors like licensing (e.g., MIT vs. proprietary vs. research-only, see Table \ref{tab:llm_metadata}) are crucial for determining practical deployability in healthcare settings.

\subsection{AI-Human Collaboration Context}
\label{subsec:ai_human_collab}

The performance of several evaluated LLMs on the PNA exam was remarkable, with many surpassing the average medical student score (101 correct) and one even exceeding the top-performing student (135 correct). While these results demonstrate significant potential, they must be interpreted cautiously. Standardized multiple-choice exams primarily assess declarative knowledge recall and specific reasoning pathways, failing to capture the full breadth of clinical competence. Crucial skills like procedural execution, nuanced patient communication, ethical decision-making under uncertainty, adapting to real-world ambiguity, and integrating complex, multimodal patient data are largely untested. Furthermore, inherent LLM limitations---including the potential for hallucination, biases inherited from training data, and inconsistent reasoning (``jagged intelligence'' \cite{karpathy_jagged_intelligence_ref})---remain significant concerns, particularly in high-stakes medical applications.

Therefore, these findings strongly reinforce the view of LLMs as powerful \textbf{assistive} tools, rather than replacements for human clinicians. This aligns with concepts like AI-Human Symbiosis, where technology augments human capabilities. LLMs can enhance clinical workflows by offering rapid information retrieval, suggesting differential diagnoses, summarizing complex literature, or drafting documentation. However, the ultimate clinical judgment, contextual understanding, empathy, and ethical responsibility must reside with the trained medical professional. Effective and safe integration hinges on leveraging AI's strengths while maintaining rigorous human oversight and critical evaluation within well-defined clinical workflows.

\begin{figure}
    \centering
    \includegraphics[width=1\linewidth]{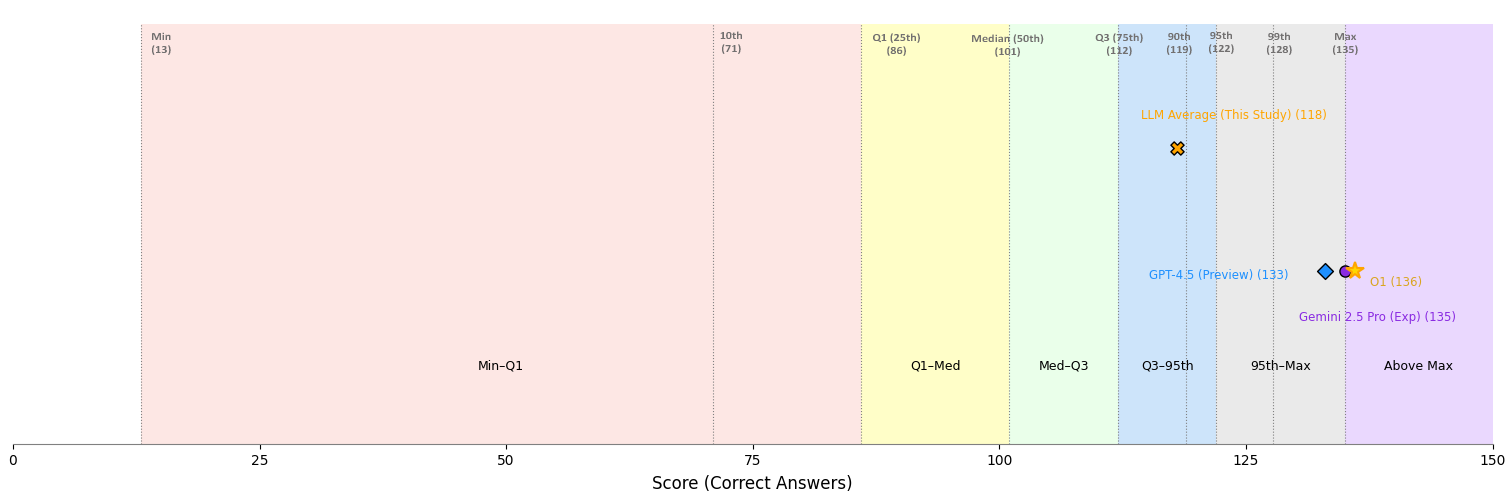}
    \caption{PNA 2024: Top 3 LLM Scores (+Avg) vs Human Percentiles Bands}
    \label{fig:me}
\end{figure}

\subsection{Optimal Model Selection Considerations}
Based on our evaluation using the PNA 2024 benchmark and the scoring metric (Eq. \ref{eq:score_final}), the choice of LLM might depend on specific priorities, as shown in Table \ref{Top3_LLMs}:
\begin{enumerate}
    \item \textbf{Peak Accuracy (Cost No Object):} \goldmodel{OpenAI O1}(136 correct) offers the highest raw accuracy, followed closely by \silvermodel{Google Gemini 2.5 Pro} (135) and \bronzemodel{OpenAI GPT-4.5 Preview} (133).
    \item \textbf{Best Overall Value (Score):} \goldmodel{Google's Gemini 2.5 Pro} (Exp, \MaxScore) and \silvermodel{Gemini 2.0 Flash Thinking} (Exp, 62.14) lead due to high accuracy combined with assumed low/zero cost, followed by \bronzemodel{Gemini 2.0 Pro (Exp 02-05)} (57.87).
    \item \textbf{Strong Performance with Moderate Cost:} \goldmodel{OpenAI O3-Mini-High} (45.79, \$3.74), \silvermodel{Grok 3 Thinking} (44.45, \$12.60), and \bronzemodel{Anthropic's Claude 3.7 Sonnet Thinking} (42.14, \$18.00) provide good balances.
    \item \textbf{Open Source:} \goldmodel{LLaMA 4 Maverick} (51.20) is a standout, outperforming other evaluated free models like \silvermodel{DeepSeek R1} (47.24) and \bronzemodel{LLaMA 3.1 405B}(Score: 41.63) in our combined score.
\end{enumerate}
\renewcommand{\arraystretch}{1.3} 
\begin{table}[h!]
\centering
\caption{Top 3 LLMs per Category}
\vspace{5pt}
\begin{tabular}{l c l}
\toprule
\textbf{Category} & \textbf{Medal} & \textbf{Model (Performance Info)} \\
\midrule

\multirow{3}{*}{\textbf{Peak Accuracy}} 
    & \GoldMedal   & \goldmodel{OpenAI O1 (136 / 150)} \\
    & \SilverMedal & \silvermodel{Google Gemini 2.5 Pro (135 / 150)} \\
    & \BronzeMedal & \bronzemodel{OpenAI GPT-4.5 Preview (133 / 150)} \\
\addlinespace
\midrule
\multirow{3}{*}{\textbf{Best Value}} 
    & \GoldMedal   & \goldmodel{Gemini 2.5 Pro (Exp, Score: 72.90)} \\
    & \SilverMedal & \silvermodel{Gemini 2.0 Flash Thinking (Score: 62.14)} \\
    & \BronzeMedal & \bronzemodel{Gemini 2.0 Pro (Exp 02-05 (Score: 57.87)} \\
\addlinespace
\midrule
\multirow{3}{*}{\textbf{Strong + Moderate Cost}}
    & \GoldMedal   & \goldmodel{OpenAI O3-Mini-High (Score: 45.79, Cost: \$3.74)} \\
    & \SilverMedal & \silvermodel{Grok 3 Thinking (Score: 44.45, Cost: \$12.60)} \\
    & \BronzeMedal & \bronzemodel{Claude 3.7 Sonnet Thinking (Score: 42.14, Cost: \$18.00)} \\
\addlinespace
\midrule
\multirow{3}{*}{\textbf{Open Source / Free}} 
    & \GoldMedal   & \goldmodel{LLaMA 4 Maverick (Score: 51.20)} \\
    & \SilverMedal & \silvermodel{DeepSeek R1 (Score: 47.24)} \\
    & \BronzeMedal & \bronzemodel{LLaMA 3.1 405B (Score: 41.63)} \\

\bottomrule
\label{Top3_LLMs}
\end{tabular}
\end{table}

Factors beyond score, such as API reliability, data privacy policies (especially for PHI), ease of integration, specific language support nuances (tested here in Portuguese), and multimodality features, would also heavily influence real-world selection for clinical deployment.
\newpage
\subsection{Regulatory Considerations: Mapping LLMs to EU AI Act}
Beyond technical performance and ethical considerations, the practical deployment of LLMs in clinical settings is increasingly governed by regulatory frameworks designed to ensure safety and trustworthiness. A landmark piece of legislation in this domain is the EU AI Act. To clarify the potential implications for developers and healthcare providers, Table~\ref{tab:eu_ai_act_map} provides a preliminary mapping of common LLM use cases in healthcare onto the risk categories defined by the Act, outlining their key consequences.
\begin{table}[h!]
\centering
\caption{Mapping common LLM use cases in healthcare to EU AI Act}
\vspace{3pt}
\footnotesize
\definecolor{rowgray}{gray}{0.96}
\definecolor{lowrisk}{RGB}{44,139,66}    
\definecolor{highrisk}{RGB}{217,133,44}  
\definecolor{prohibited}{RGB}{198,53,53} 
\rowcolors{2}{rowgray}{} 
\label{tab:eu_ai_act_map}
\begin{tabular}{|p{3.4cm}|p{3.1cm}|p{7.3cm}|}
\rowcolor{gray!23}
\hline
\textbf{LLM Use Case} & \textbf{EU AI Act Risk Class} & \textbf{Key Implications} \\
\hline
Chatbot for patient FAQs                & \textcolor{lowrisk}{Low / Minimal}  & Exempt from medical device rules. General disclosure recommended; minimal oversight.\\
Clinical decision support (diagnosis/treatment suggestion) & \textcolor{highrisk}{High}   & Classified as medical device: requires transparency, risk management, documentation, and mandatory human oversight.\\
Automated diagnosis/treatment (no human in loop) & \textcolor{prohibited}{Prohibited / Very High} & Banned or not permitted unless strictly mediated by a physician; must offer full traceability and audits.\\
LLM for research or education only      & \textcolor{lowrisk}{Low}            & Simple disclosure of AI use; no extra compliance for clinical safety.\\
\hline
\end{tabular}
\end{table}
\newpage
\section{Future Directions}
\subsection{Limitations}
\label{subsec:limitations}
This study has several limitations that warrant consideration:
\begin{itemize}
    \item The evaluation used a single multiple-choice exam (PNA 2024) focused on broad medical knowledge. This may not fully represent the complexities of real-time, interactive clinical diagnosis and treatment planning, nor specific specialized domains.
    \item The exam was conducted in Portuguese, evaluating models on their proficiency in this specific language context, which might differ from their English performance or performance in other languages.
    \item Performance was measured using basic prompting and a strict pass@1 methodology (first attempt only). Results could potentially differ with more sophisticated prompt engineering or if multiple attempts/interaction were allowed.
    \item Cost estimations are approximate, based on publicly available data at a specific point in time (Q1 2025), and subject to change. Pricing for experimental models is uncertain.
    \item The study did not include systematic qualitative error analysis to understand the nature of mistakes (e.g., reasoning errors vs. knowledge gaps vs. hallucinations).
\end{itemize}

\subsection{Future Work}
Future work should focus on:
\begin{enumerate}
    \item \textbf{Clinical Vignette Evaluation:} Testing models on realistic clinical case scenarios requiring differential diagnosis, treatment planning, and justification, potentially in interactive settings.
    \item \textbf{Specialty-Specific Testing:} Assessing performance in specific medical domains (e.g., radiology report interpretation, adherence to cardiology guidelines, oncology treatment options).
    \item \textbf{Real-World Integration Studies:} Evaluating the impact of LLM assistance in actual clinical workflows, measuring effects on diagnostic time/accuracy, clinician satisfaction, and potentially patient outcomes.
    \item \textbf{Performance Comparison with Clinicians:} Directly comparing LLM performance (accuracy, reasoning quality, safety) against experienced medical professionals on defined diagnostic or treatment tasks.
    \item \textbf{Safety, Reliability and Bias:} Developing robust methods for validating outputs, detecting hallucinations/errors, ensuring patient safety, and rigorously assessing and mitigating biases across diverse patient populations. This includes qualitative error analysis.
    \item \textbf{Investigating Novel Architectures:} Exploring techniques like diffusion models adapted for text \cite{b6} for specific medical generation tasks (e.g., summarization).
    \item \textbf{Ethical and Regulatory Frameworks:} Continuing collaborative efforts to develop clear guidelines and standards for responsible LLM deployment in healthcare, compliant with regulations like GDPR and the EU AI Act.
\end{enumerate}
These steps are crucial for responsibly translating the demonstrated potential of LLMs into tangible benefits for patient care.
\subsection{A True, Blinded Benchmark: 2025 Live PNA LLM Challenge with Anonymized "Students"}
To eliminate even the \emph{possibility} of data leakage and contamination, we strongly advocate a \textbf{live, prospective benchmark} for 2025: 

\begin{sidebox}[orange!12]{\textbf{Proposal: PNA 2025 LLM-Student Showdown}}
\small
On exam day, present the exact 2025 PNA to the top-5 LLMs \textbf{simultaneously} with real students. Each model’s responses should be attributed to a \emph{fictitious “student”} (e.g., ``Student Alpha'', ``Student Beta''), without any indication of whether an answer came from a model or human. Compare LLM performance directly with anonymized human results. This design:
\begin{itemize}\itemsep2pt
    \item Guarantees no model has seen or trained on the questions;
    \item Allows fully fair, blinded analysis;
    \item Will establish the most trustworthy LLM test-to-date for clinical readiness.
\end{itemize}
\end{sidebox}

\section{Conclusion}
\label{sec:conclusion}

Large Language Models demonstrate significant and rapidly growing capabilities relevant to supporting medical diagnosis and treatment. Our evaluation of 21 contemporary LLMs on the challenging 2024 Portuguese National Exam for medical specialty access reveals substantial potential, with several models outperforming human student benchmarks in terms of accuracy. Crucially, our analysis highlights that some of the best-performing models, when considering both accuracy and cost-effectiveness using our defined metric, are available at low or potentially no cost (including high-performing open-source options like DeepSeek R1 and promising experimental models from Google). This suggests that powerful AI assistance could become broadly accessible across diverse healthcare settings. Models employing structured reasoning techniques like Chain-of-Thought show particular promise for complex tasks requiring explainable outputs, though efficiency-focused variants like Chain-of-Draft warrant further investigation in clinical contexts.

However, significant variations in performance, cost, licensing, and other characteristics exist across models and providers. Furthermore, the responsible deployment of these tools in clinical practice necessitates careful consideration and mitigation of challenges related to accuracy, reliability (including the risk of hallucination and "jagged intelligence"), data privacy, workflow integration, bias, ethical use, and regulatory compliance (such as GDPR and the EU AI Act). The results strongly reinforce the view that LLMs are best positioned as powerful complementary tools to augment, not replace, the expertise, critical judgment, and holistic care provided by human healthcare professionals. Continued research, rigorous validation in realistic clinical settings, systematic error analysis, and the development of clear implementation guidelines are essential to harness the potential of LLMs safely and effectively to improve patient care and outcomes.

\section*{Acknowledgments}
The authors acknowledge the support of the AITriage4SU Project (2024.07400.IACDC/2024), funded by the FCT (Foundation for Science and Technology), Portugal, which made this research possible.

\bibliographystyle{unsrt}
\bibliography{references}

\begin{thebibliography}{10}

\bibitem{b1}
{John Snow Labs}.
\newblock {LLMs} taking on medical challenge problems.
\newblock \url{https://www.johnsnowlabs.com/llms-taking-on-medical-challenge-problems/}.
\newblock Accessed on: Apr. 10, 2025.

\bibitem{b2}
G.~Ferraz-Costa, M.~Grin{\'e}, M.~Oliveira-Santos, and R.~Teixeira.
\newblock Performance of {ChatGPT} in the {P}ortuguese {N}ational {R}esidency {A}ccess {E}xamination.
\newblock {\em Acta Medica Portuguesa}, 38(3):170--174, March 2025.

\bibitem{b4}
M.~Nori, E.~King, S.~McKinney, H.~Carignan, and E.~Horvitz.
\newblock Capabilities of {GPT-4} on medical challenge problems.
\newblock {\em arXiv preprint arXiv:2303.13375}, 2023.

\bibitem{yang2023gptsolvemathematicalproblems}
Zhen Yang, Ming Ding, Qingsong Lv, Zhihuan Jiang, Zehai He, Yuyi Guo, Jinfeng Bai, and Jie Tang.
\newblock {GPT} can solve mathematical problems without a calculator.
\newblock {\em arXiv eprint arXiv:2309.03241}, 2023.

\bibitem{b5}
S.~Xu, W.~Xie, L.~Zhao, and P.~He.
\newblock Chain of draft: Thinking faster by writing less.
\newblock {\em arXiv preprint arXiv:2502.18600}, 2025.

\bibitem{diffusion_survey_ref}
F.~A. Croitoru et~al.
\newblock Diffusion models in vision: {A} survey.
\newblock {\em IEEE Transactions on Pattern Analysis and Machine Intelligence}, 45(9):10800--10820, September 2023.

\bibitem{b6}
Q.~Yi, X.~Chen, C.~Zhang, Z.~Zhou, L.~Zhu, and X.~Kong.
\newblock Diffusion models in text generation: a survey.
\newblock {\em PeerJ Computer Science}, 10:e1905, 2024.

\bibitem{inception_mercury_ref}
{Inception Labs}.
\newblock Introducing {M}ercury, the first commercial-scale diffusion large language model.
\newblock Medium, February 2025.
\newblock \url{https://medium.com/@aajaysaikiran/introducing-mercury-the-first-commercial-scale-diffusion-large-language-model-703dd6e67b81}. Accessed on: Apr. 10, 2025.

\bibitem{inception_website_ref}
{Inception Labs}.
\newblock Introducing a new generation of language models.
\newblock \url{https://inceptionlabs.ai/}.
\newblock Accessed on: Apr. 10, 2025.

\bibitem{karpathy_jagged_intelligence_ref}
A.~Karpathy.
\newblock State of {GPT}.
\newblock Presented at Microsoft Build Conference. [Online Video], May 2023.
\newblock Available: \url{https://www.youtube.com/watch?v=bZQun8Y4L2A}. Accessed on: Apr. 10, 2025.

\bibitem{lightman2023letsverifystepstep}
Hunter Lightman, Vineet Kosaraju, Yura Burda, Harri Edwards, Bowen Baker, Teddy Lee, Jan Leike, John Schulman, Ilya Sutskever, and Karl Cobbe.
\newblock Let's verify step by step, 2023.

\bibitem{Kirchner2024ProverVerifierGames}
Jan~Hendrik Kirchner, Yining Chen, Harri Edwards, Jan Leike, Nat McAleese, and Yuri Burda.
\newblock Prover-verifier games improve legibility of llm outputs, 2024.

\bibitem{lecun_fridman_2023}
Yann LeCun.
\newblock {Yann LeCun: AI, Consciousness, and the Future of Intelligence}.
\newblock Lex Fridman Podcast \#364, February 2023.
\newblock Discussion on AI as tools/amplifiers/assistants for humans, rather than direct replacements, is a recurring theme throughout the conversation, particularly when discussing AGI and future AI systems. Accessed on: Apr. 14, 2025.

\bibitem{alphafold2021}
John Jumper et~al.
\newblock Highly accurate protein structure prediction with {AlphaFold}.
\newblock {\em Nature}, 596(7873):583--589, August 2021.

\bibitem{ch_2025_nobel_prize}
{Royal Swedish Academy of Sciences}.
\newblock Press release: The nobel prize in chemistry 2024.
\newblock Nobel Prize Official Website, October 2024.
\newblock Available: \url{https://www.nobelprize.org/prizes/chemistry/2024/press-release/}. Accessed on: Apr. 10, 2025.

\bibitem{pna2024}
{Administra\c{c}\~{a}o Central do Sistema de Sa\'{u}de (ACSS)}.
\newblock Prova nacional de acesso 2024.
\newblock Website Oficial da ACSS, Sec\c{c}\~{a}o Internato M\'{e}dico/PNA, 2024.
\newblock Available: \url{https://www.acss.min-saude.pt/category/internato-medico/pna/}. Accessed on: Apr. 10, 2025.

\bibitem{b3}
J.~M. Hoppe, M.~K. Auer, A.~Strüven, S.~Massberg, and C.~Stremmel.
\newblock {ChatGPT} with {GPT-4} outperforms emergency department physicians in diagnostic accuracy: Retrospective analysis.
\newblock {\em Journal of Medical Internet Research}, 26:e56110, 2024.

\bibitem{GoogleModelsVertexAI}
{Google Cloud}.
\newblock Generative {AI} models.
\newblock Google Cloud Documentation, 2024.
\newblock Available: \url{https://cloud.google.com/vertex-ai/generative-ai/docs/learn/models}. Accessed on: Apr. 10, 2025.

\bibitem{ValsAI_Gemini2FlashThink}
{Vals.ai}.
\newblock Google {G}emini 2.0 {F}lash {T}hinking {E}xp 01-21 model card.
\newblock Vals.ai, 2025.
\newblock Available: \url{https://www.vals.ai/models/google_gemini-2.0-flash-thinking-exp-01-21}. Accessed on: Apr. 10, 2025.

\bibitem{ValsAI_Gemini2ProExp}
{Vals.ai}.
\newblock Google {G}emini 2.0 {P}ro {E}xp 02-05 model card.
\newblock Vals.ai, 2025.
\newblock Available: \url{https://www.vals.ai/models/google_gemini-2.0-pro-exp-02-05}. Accessed on: Apr. 10, 2025.

\bibitem{GeminiAPIModels}
{Google AI for Developers}.
\newblock Gemini models.
\newblock Google AI Documentation, 2025.
\newblock Available: \url{https://ai.google.dev/gemini-api/docs/models}. Accessed on: Apr. 10, 2025.

\bibitem{Gemma3_27B_OpenRouter}
{OpenRouter}.
\newblock Gemma 3 27b.
\newblock OpenRouter Models.
\newblock Available: \url{https://openrouter.ai/google/gemma-3-27b-it}. Accessed on: Apr. 10, 2025.

\bibitem{Gemma3Blog}
{Google Developers Blog}.
\newblock Gemma 3: {G}oogle’s new open model family.
\newblock Google Blog, March 2025.
\newblock Available: \url{https://blog.google/technology/developers/gemma-3/}. Accessed on: Apr. 10, 2025.

\bibitem{OpenAIReleaseNotes}
{OpenAI Help Center}.
\newblock Model release notes.
\newblock OpenAI Support.
\newblock Available: \url{https://help.openai.com/en/articles/9624314-model-release-notes}. Accessed on: Apr. 10, 2025.

\bibitem{GPT4oWikipedia}
{Wikipedia contributors}.
\newblock {GPT-4o}.
\newblock Wikipedia, The Free Encyclopedia.
\newblock Available: \url{https://en.wikipedia.org/wiki/GPT-4o}. Accessed on: Apr. 10, 2025.

\bibitem{OpenAIHelloGPT4o}
{OpenAI}.
\newblock Hello {GPT-4o}.
\newblock OpenAI Blog, May 2024.
\newblock Available: \url{https://openai.com/index/hello-gpt-4o/}. Accessed on: Apr. 10, 2025.

\bibitem{GPT4oKnowledgeCutoffBGR}
{BGR}.
\newblock {GPT-4o} is the best {ChatGPT} model, but its knowledge cutoff is worse than {ChatGPT’s} {GPT-4}.
\newblock BGR Tech.
\newblock Available: \url{https://bgr.com/tech/gpt-4o-knowledge-cutoff-is-worse-than-chatgpts-gpt-4/}. Accessed on: Apr. 10, 2025.

\bibitem{OpenAI_O3Mini}
{OpenAI}.
\newblock Introducing {O}pen{AI} o3-mini.
\newblock OpenAI Blog, January 2025.
\newblock Available: \url{https://openai.com/index/openai-o3-mini/}. Accessed on: Apr. 10, 2025.

\bibitem{OpenRouterO3Mini}
{OpenRouter}.
\newblock {OpenAI} o3-mini-high.
\newblock OpenRouter Models.
\newblock Available: \url{https://openrouter.ai/openai/o3-mini-high}. Accessed on: Apr. 10, 2025.

\bibitem{Claude37SonnetSystemCard}
{Anthropic}.
\newblock Claude 3.7 {S}onnet system card.
\newblock Anthropic.
\newblock Available: \url{https://anthropic.com/claude-3-7-sonnet-system-card}. Accessed on: Apr. 10, 2025.

\bibitem{Claude37SonnetAWSBlog}
{AWS News Blog}.
\newblock {Anthropic’s} {Claude} 3.7 {S}onnet, the first hybrid reasoning model, is now available in {A}mazon {B}edrock.
\newblock AWS Blog, February 2025.
\newblock Available: \url{https://aws.amazon.com/blogs/aws/anthropics-claude-3-7-sonnet-the-first-hybrid-reasoning-model-is-now-available-in-amazon-bedrock/}. Accessed on: Apr. 10, 2025.

\bibitem{Claude35SonnetNews}
{Anthropic}.
\newblock Introducing {Claude} 3.5 {S}onnet.
\newblock Anthropic News, June 2024.
\newblock Available: \url{https://www.anthropic.com/news/claude-3-5-sonnet}. Accessed on: Apr. 10, 2025.

\bibitem{ClaudeUpgradeOct24}
{Anthropic}.
\newblock Updates to the {Claude} 3 family.
\newblock Anthropic News, October 2024.
\newblock Available: \url{https://www.anthropic.com/news/updates-to-the-claude-3-family}. Accessed on: Apr. 10, 2025.

\bibitem{Claude35HaikuAWSBlog}
{AWS What's New}.
\newblock {Anthropic’s} {Claude} 3.5 {H}aiku model is now available in {A}mazon {B}edrock.
\newblock AWS, November 2024.
\newblock Available: \url{https://aws.amazon.com/about-aws/whats-new/2024/11/anthropics-claude-3-5-haiku-model-amazon-bedrock/}. Accessed on: Apr. 10, 2025.

\bibitem{Claude35HaikuNews}
{Anthropic}.
\newblock Claude 3.5 {H}aiku.
\newblock Anthropic.
\newblock Available: \url{https://www.anthropic.com/claude/haiku}. Accessed on: Apr. 10, 2025.

\bibitem{DeepSeekR1Release}
{DeepSeek AI}.
\newblock {DeepSeek-R1} release.
\newblock DeepSeek API Docs, January 2025.
\newblock Available: \url{https://api-docs.deepseek.com/news/news250120}. Accessed on: Apr. 10, 2025.

\bibitem{DeepSeekR1HF}
{Hugging Face}.
\newblock deepseek-ai/{DeepSeek-R1} model card.
\newblock Hugging Face Hub.
\newblock Available: \url{https://huggingface.co/deepseek-ai/DeepSeek-R1}. Accessed on: Apr. 10, 2025.

\bibitem{DeepSeekV3InfoQ}
{InfoQ}.
\newblock {DeepSeek} open-sources {DeepSeek-V3} {LLM} family.
\newblock InfoQ, January 2025.
\newblock Available: \url{https://www.infoq.com/news/2025/01/deepseek-v3-llm/}. Accessed on: Apr. 10, 2025.

\bibitem{DeepSeekV3HPCWire}
{HPCwire}.
\newblock {DeepSeek-V3-0324} quietly lands on {H}ugging {F}ace.
\newblock HPCwire.
\newblock Available: \url{https://www.hpcwire.com/off-the-wire/deepseek-v3-0324-quietly-lands-on-hugging-face/}. Accessed on: Apr. 10, 2025.

\bibitem{DeepSeekV3HF}
{Hugging Face}.
\newblock deepseek-ai/{DeepSeek-V3-Chat} model card.
\newblock Hugging Face Hub.
\newblock Available: \url{https://huggingface.co/deepseek-ai/DeepSeek-V3-Chat}. Accessed on: Apr. 10, 2025.

\bibitem{Llama4ReleaseBlog}
{Meta AI}.
\newblock Introducing {L}lama 4.
\newblock Meta AI Blog, April 2025.
\newblock [URL Needed]. Accessed on: Apr. 10, 2025.

\bibitem{Llama4HF}
{Hugging Face}.
\newblock meta-llama/{Meta-Llama-4-XXXB-Instruct}.
\newblock Hugging Face Hub.
\newblock [URL Needed]. Accessed on: Apr. 10, 2025.

\bibitem{LLaMA33_70B_HF}
{Hugging Face}.
\newblock meta-llama/{Llama-3.3-70B-Instruct}.
\newblock Hugging Face Hub.
\newblock Available: \url{https://huggingface.co/meta-llama/Llama-3.3-70B-Instruct}. Accessed on: Apr. 10, 2025.

\bibitem{LLaMA33_70B_AWS}
{AWS What's New}.
\newblock {Meta’s} {L}lama 3.3 70{B} model is now available in {A}mazon {B}edrock.
\newblock AWS, December 2024.
\newblock Available: \url{https://aws.amazon.com/about-aws/whats-new/2024/12/metas-llama-3-3-70b-model-amazon-bedrock/}. Accessed on: Apr. 10, 2025.

\bibitem{LLaMA31_405B_AWS}
{AWS What's New}.
\newblock {Meta} {L}lama 3.1 405{B} now generally available on {A}mazon {B}edrock.
\newblock AWS, July 2024.
\newblock Available: \url{https://aws.amazon.com/about-aws/whats-new/2024/07/meta-llama-3-1-405b-generally-available-amazon-bedrock/}. Accessed on: Apr. 10, 2025.

\bibitem{Llama31Release}
{Meta AI}.
\newblock Introducing {L}lama 3.1.
\newblock Meta AI Blog, July 2024.
\newblock Available: \url{https://ai.meta.com/blog/meta-llama-3-1/}. Accessed on: Apr. 10, 2025.

\bibitem{Grok3Verge}
{The Verge}.
\newblock {E}lon {M}usk’s {G}rok-3 {AI} beats {ChatGPT-4} in benchmarks.
\newblock The Verge Command Line, February 2025.
\newblock Available: \url{https://www.theverge.com/command-line-newsletter/617780/grok-3-elon-musk-ai-race-chatgpt}. Accessed on: Apr. 10, 2025.

\bibitem{Grok3Helicone}
{Helicone Blog}.
\newblock {G}rok 3 technical review \& benchmark comparison.
\newblock Helicone.
\newblock Available: \url{https://www.helicone.ai/blog/grok-3-benchmark-comparison}. Accessed on: Apr. 10, 2025.

\bibitem{QwenWikipedia}
{Wikipedia contributors}.
\newblock Qwen.
\newblock Wikipedia, The Free Encyclopedia.
\newblock Available: \url{https://en.wikipedia.org/wiki/Qwen}. Accessed on: Apr. 10, 2025.

\bibitem{QwenLLMAlibabaDoc}
{Alibaba Cloud Documentation}.
\newblock Qwen {LLM}s.
\newblock Alibaba Cloud.
\newblock Available: \url{https://www.alibabacloud.com/help/en/model-studio/developer-reference/what-is-qwen-llm}. Accessed on: Apr. 10, 2025.

\bibitem{Qwen25MaxBlog}
{Qwen Team Blog}.
\newblock {Qwen2.5-Max}: Exploring the intelligence boundary of {M}o{E} {LLM}.
\newblock QwenLM GitHub Blog.
\newblock Available: \url{https://qwenlm.github.io/blog/qwen2.5-max/}. Accessed on: Apr. 10, 2025.

\bibitem{MistralLarge2411HF_TechxGenus}
{Hugging Face}.
\newblock {TechxGenus}/{Mistral-Large-Instruct-2411-AWQ}.
\newblock Hugging Face Hub.
\newblock Available: \url{https://huggingface.co/TechxGenus/Mistral-Large-Instruct-2411-AWQ}. Accessed on: Apr. 10, 2025.

\bibitem{MistralLarge2411MSBlog}
{Microsoft Community Hub}.
\newblock Introducing {M}istral {L}arge 2411: Transforming industries with cutting-edge {AI} capabilities.
\newblock Microsoft Tech Community, November 2024.
\newblock Available: \url{https://techcommunity.microsoft.com/blog/machinelearningblog/introducing-mistral-large-2411-transforming-industries-with-cutting-edge-ai-capa/4357414}. Accessed on: Apr. 10, 2025.

\bibitem{MistralSmall31News}
{Mistral AI}.
\newblock Mistral {S}mall 3.1.
\newblock Mistral AI News, March 2025.
\newblock Available: \url{https://mistral.ai/news/mistral-small-3-1/}. Accessed on: Apr. 10, 2025.

\bibitem{MistralSmall31_HF}
{Hugging Face}.
\newblock mistralai/{Mistral-Small-3.1-24B-Base-2503}.
\newblock Hugging Face Hub.
\newblock Available: \url{https://huggingface.co/mistralai/Mistral-Small-3.1-24B-Base-2503}. Accessed on: Apr. 10, 2025.

\end{thebibliography}



\newpage 
\begin{landscape}
\appendix 
\setcounter{table}{0} 
\renewcommand{\thetable}{A.\arabic{table}} 


\centering
\renewcommand{\arraystretch}{1.3}
\setlength{\tabcolsep}{5pt} 
\scriptsize
\rowcolors{2}{gray!10}{white} 

\begin{longtable}{@{}>{\raggedright\arraybackslash}p{3.8cm} p{1.5cm} p{1.5cm} p{1.6cm} p{2.2cm} p{1.4cm} p{3.4cm} p{3.4cm}@{}}

\caption{\textbf{Large Language Models Used in PNA Evaluation Study: Metadata Overview}\\ 
\small(Grouped by Developer; refer to bibliography for sources. Information as of Apr 10, 2025.)}
\label{tab:llm_metadata} \\

\toprule
\textbf{Model Name} & \textbf{Developer} & \textbf{Knowledge Cutoff} & \textbf{Release Date} & \textbf{License/Availability} & \textbf{Context Window} & \textbf{Architecture \& Training Notes} & \textbf{Sources} \\
\midrule
\endfirsthead

\multicolumn{8}{c}%
{{\bfseries \tablename\ \thetable{} -- continued from previous page}} \\
\toprule
\textbf{Model Name} & \textbf{Developer} & \textbf{Knowledge Cutoff} & \textbf{Release Date} & \textbf{License/Availability} & \textbf{Context Window} & \textbf{Architecture \& Training Notes} & \textbf{Sources} \\
\midrule
\endhead

\multicolumn{8}{r}{{Continued on next page}} \\
\endfoot

\bottomrule 
\endlastfoot


\multicolumn{8}{l}{\cellcolor{googleblue}\textbf{Google (DeepMind)}} \\
Gemini 2.5 Pro (Exp) & Google & Jan 2025 & Mar 25, 2025 & Closed; Vertex AI Preview & 1M tokens & Top-tier Gemini 2.5, multimodal (text, image, video, audio). & \cite{GoogleModelsVertexAI} \\
Gemini 2.0 Flash Thinking (Exp) & Google & May 2024 & Jan 22, 2025 & Closed; API (Vertex AI) & 1M/64k tok. (Adv./Out) & Multimodal Gemini 2.0 tech, opt. for CoT. & \cite{GoogleModelsVertexAI, ValsAI_Gemini2FlashThink} \\
Gemini 2.0 Pro (Exp 02‑05) & Google & Jan 2025 & Feb 5, 2025 & Closed; Experimental API & 2M tokens & Advanced Gemini 2.0 Pro variant, multimodal, long context. & \cite{ValsAI_Gemini2ProExp, GeminiAPIModels} \\
Gemma 3 27B & Google & Aug 2024 & Mar 12, 2025 & Open-source (Gemma lic.) & 128k tokens & 27B open model (Gemini 2.0 based), multimodal (image, video). & \cite{Gemma3_27B_OpenRouter, Gemma3Blog} \\
\midrule 

\multicolumn{8}{l}{\cellcolor{openaigreen}\textbf{OpenAI}} \\
OpenAI GPT-4.5 (Preview) & OpenAI & Oct 2023 & Fev 27, 2025 & Closed; API/ChatGPT & 128k tokens & Enhanced multimodal model, improved reasoning and context; preview/research release, successor to GPT-4o. & \cite{OpenAIReleaseNotes, GPT4oWikipedia} \\
OpenAI GPT-4o & OpenAI & Oct 2023 (Base) / Nov 2024 (ChatGPT) & Nov 20, 2024 & Closed; API/ChatGPT & 128k tokens & Flagship multimodal (text, image, audio), successor to GPT-4. & \cite{GPT4oWikipedia, OpenAIHelloGPT4o, GPT4oKnowledgeCutoffBGR} \\
OpenAI O1 & OpenAI & ~Oct 2023 & Dec 17, 2024 & Closed; Research preview API & ~128k tokens (est.) & Experimental reasoning model, successor to O3-Mini. & \cite{OpenAIReleaseNotes} \\
OpenAI O3-Mini (High mode) & OpenAI & ~Oct 2023 & Jan 31, 2025 & Closed; API/ChatGPT & 200k tok. (input) & Small efficient reasoning model (like GPT-4o Mini), RL, CoT output. & \cite{OpenAI_O3Mini, OpenRouterO3Mini} \\
OpenAI GPT-4o Mini & OpenAI & Oct 2023 & Jul 18, 2024 & Closed; ChatGPT/API (cheaper tier) & 128k tokens & Scaled-down GPT-4o, cost-efficient multimodal capabilities. & \cite{GPT4oWikipedia, OpenAIHelloGPT4o} \\
\midrule 

\multicolumn{8}{l}{\cellcolor{anthropicpurple}\textbf{Anthropic}} \\
Claude 3.7 Sonnet (Thinking) & Anthropic & Nov 2024 & Feb 24, 2025 & Closed; API (Claude.ai, Bedrock) & 200k tokens & Advanced hybrid reasoning, self-reflects (CoT), multimodal. Long code output support. & \cite{Claude37SonnetSystemCard, Claude37SonnetAWSBlog} \\
Claude 3.5 Sonnet (New) & Anthropic & Apr 2024 & Oct 22, 2024 & Closed; API/Claude.ai & 200k tokens & Strong vision/language, faster than Opus 3. Upgraded w/ tool use (Oct '24). & \cite{Claude35SonnetNews, ClaudeUpgradeOct24} \\
Claude 3.5 Haiku & Anthropic & July 2024 & Nov 4, 2024 & Closed; API (Instant tier) & 200k tokens & Fast, lightweight model for high-speed responses, improved tool use. & \cite{Claude35HaikuAWSBlog, Claude35HaikuNews} \\
\midrule 

\multicolumn{8}{l}{\cellcolor{deepseekorange}\textbf{DeepSeek AI}} \\
DeepSeek R1 & DeepSeek AI & Jan 2025 & Jan 20, 2025 & Open-source (MIT) & 8k+ (ext.) & 685B MoE (total params?), RL for CoT generation. & \cite{DeepSeekR1Release, DeepSeekR1HF} \\
DeepSeek V3 & DeepSeek AI & Dec 2024 & Mar 24, 2025 & Open-source (MIT) & Large context (32k default) & 671B MoE (256 experts), open-weight, FP8 precision. & \cite{DeepSeekV3InfoQ, DeepSeekV3HPCWire, DeepSeekV3HF} \\
\midrule 

\multicolumn{8}{l}{\cellcolor{metared}\textbf{Meta AI}} \\
Llama 4 Maverick & Meta AI & Aug 2024 & Apr 5, 2025 & Source-available (Llama 4 Community Lic.) & 1M tokens & MoE (17B active / ~400B total / 128 experts), natively multimodal (text, image). & \cite{Llama4ReleaseBlog, Llama4HF} \\
Llama 3.3 70B (Instruct) & Meta AI & Dec 2023 & Dec 6, 2024 & Source-available (Llama 3.3 Community Lic.) & 128k tokens & 70B instruction-tuned, multilingual, training improvements. & \cite{LLaMA33_70B_HF, LLaMA33_70B_AWS} \\
Llama 3.1 405B & Meta AI & Dec 2023 & Jul 23, 2024 & Source-available (Llama 3.1 Lic.; research focus) & 128k tokens & 405B dense model, improved multilingual dialogue \& reasoning. & \cite{LLaMA31_405B_AWS, Llama31Release} \\
\midrule 

\multicolumn{8}{l}{\cellcolor{xaicyan}\textbf{xAI}} \\
Grok 3 (Thinking Mode) & xAI & Nov 2024 & Apr 9, 2025 & Closed; via X platform & Long CoT output & Large model with explicit "Thinking" (CoT) mode, integrates X search. & \cite{Grok3Verge, Grok3Helicone} \\
\midrule 

\multicolumn{8}{l}{\cellcolor{alibabayellow}\textbf{Alibaba Cloud}} \\
Qwen 2.5-Max & Alibaba Cloud & ~Dec 2023 & Feb 1, 2025 & Closed (API via Alibaba Cloud) & 32k/8k tok. (in/out est.) & Massive MoE (>100B), SOTA Chinese/English benchmarks. & \cite{QwenWikipedia, QwenLLMAlibabaDoc, Qwen25MaxBlog} \\
\midrule 

\multicolumn{8}{l}{\cellcolor{mistralteal}\textbf{Mistral AI}} \\
Mistral Large 2 (24.11) & Mistral AI & ~2024 (multimodal) & Nov 18, 2024 & Research-only (Mistral Research Lic.) & 128k tokens & Frontier multimodal MoE (~400B total params), long context. & \cite{MistralLarge2411HF_TechxGenus, MistralLarge2411MSBlog} \\
Mistral Small 3.1 24B & Mistral AI & ~2023 (multi-domain) & Mar 17, 2025 & Open-source (Apache 2.0) & 128k tokens & 24B multimodal (image support), long context, efficient (runs on single 24GB GPU). & \cite{MistralSmall31News, MistralSmall31_HF} \\

\end{longtable}

\vspace{0.2cm}
\begin{minipage}{0.9\linewidth} 
    \centering 
    \tiny 
    \textbf{Notes:} Context Window indicates maximum tokens unless specified otherwise (e.g., input/output). License/Availability describes primary access method. Training details based on public announcements. M=Million, k=thousand, B=Billion params. CoT=Chain-of-Thought. MoE=Mixture-of-Experts (total params often reported; active may be smaller). RL=Reinforcement Learning. SOTA=State-of-the-Art. est.=estimated. lic.=License. mod.=modification. opt.=optimized. ltd=limited. restrict.=restriction. MAU=Monthly Active Users. Exp=Experimental. Adv.=Advanced. tok.=tokens.\\ 
    \textbf{Colors represent developer groups.} Data verified as of Apr 10, 2025.
\end{minipage}

\end{landscape}

\setcounter{figure}{0} 
\renewcommand{\thefigure}{A.\arabic{figure}} 
\begin{landscape}
\begin{figure*}[ht]
    \centering
    \includegraphics[width=1.4\textwidth]{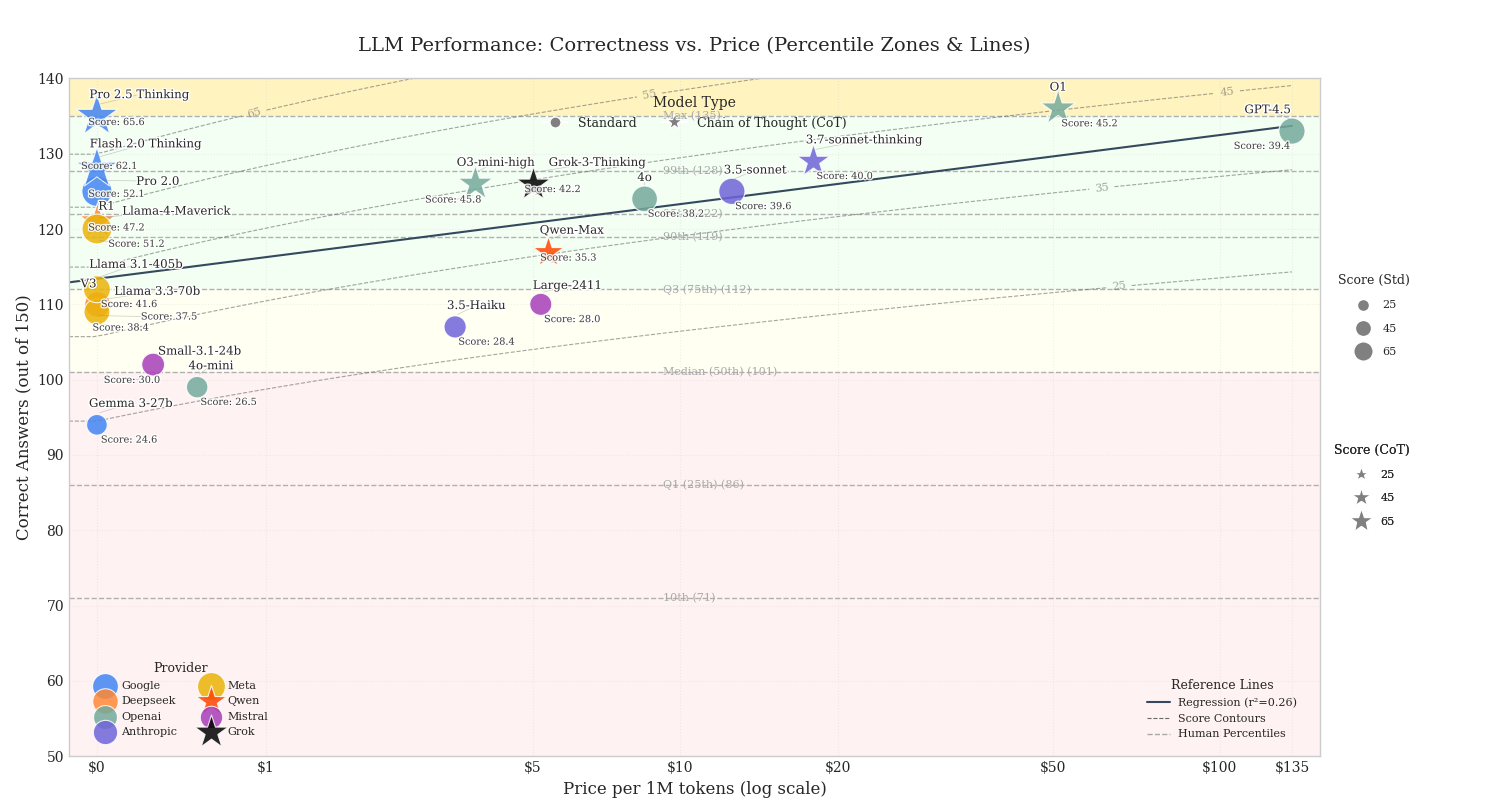} 
    \caption{\textbf{LLM Performance on PNA 2024: Accuracy vs. Estimated Cost per Million Tokens}. Bubble size is proportional to the calculated Score (Eq. \ref{eq:score_final}), emphasizing models with high accuracy and low cost.} 
    \label{fig:llm_performance_chart_appendix} 
\end{figure*}
\end{landscape}

\end{document}